\documentclass[10pt,twocolumn,letterpaper]{article}

\usepackage{cvpr}
\usepackage{times}
\usepackage{epsfig}
\usepackage{graphicx}
\usepackage{amsmath}
\usepackage{amssymb}
\usepackage{setspace}
\usepackage{array}
\usepackage{amsmath}
\usepackage{cases}
\usepackage{bbm}
\usepackage{caption}
\usepackage{subcaption}
\usepackage{graphicx}
\usepackage{threeparttable}

\usepackage[colorlinks,urlcolor=blue]{hyperref}

\cvprfinalcopy 

\def\cvprPaperID{****} 

\begin{document}

\title{EmbedMask: Embedding Coupling for One-stage Instance Segmentation}


\author{Hui Ying$^{\dag}$\thanks{The work was done when Hui Ying and Zhaojin Huang were interns in Tencent Youtu.}\ \ \ \ \ Zhaojin Huang$^{\ddag}$\footnotemark[1]\ \ \ \ \ Shu Liu$^{\S}$\ \ \ \ Tianjia Shao$^{\dag}$\ \ \ \ Kun Zhou$^{\dag}$ \\
$^{\dag}$ State Key Lab of CAD\&CG, Zhejiang University \\ 
$^{\ddag}$Institute of AI, School of EIC, Huazhong University of Science and Technology \\
$^{\S}$Tencent Youtu Lab \\
{\tt\small huiying@zju.edu.cn zhaojinhuang@hust.edu.cn \{liushuhust, tianjiashao\}@gmail.com kunzhou@acm.org}
}

\maketitle

\begin{abstract}
Current instance segmentation methods can be categorized into segmentation-based methods that segment first then do clustering, and proposal-based methods that detect first then predict masks for each instance proposal using repooling. In this work, we propose a one-stage method, named EmbedMask, that unifies both methods by taking advantages of them. Like proposal-based methods, EmbedMask builds on top of detection models making it strong in detection capability. Meanwhile, EmbedMask applies extra embedding modules to generate embeddings for pixels and proposals, where pixel embeddings are guided by proposal embeddings if they belong to the same instance. Through this embedding coupling process, pixels are assigned to the mask of the proposal if their embeddings are similar. The pixel-level clustering enables EmbedMask to generate high-resolution masks without missing details from repooling, and the existence of proposal embedding simplifies and strengthens the clustering procedure to achieve high speed with higher performance than segmentation-based methods. Without any bells and whistles, EmbedMask achieves comparable performance as Mask R-CNN, which is the representative two-stage method, and can produce more detailed masks at a higher speed. 
Code is available at \href{https://github.com/yinghdb/EmbedMask}{github.com/yinghdb/EmbedMask}


\end{abstract}

\section{Introduction}

In light of the rapid development of deep learning and machine industry, a lot of tasks ~\cite{girshick2014rich, he2017mask, he2016deep, krizhevsky2012imagenet, liu2016ssd, redmon2016youyolo, ren2015faster,zhao2017pyramidpspnet} in the field of computer vision have made tremendous progress.
We can also observe that the application of deep networks in computer vision has extended from image-level to pixel-level. Specifically, instance segmentation can be viewed as an extension of object detection, which extends the detected objects from instance-level to pixel-level. 

\begin{figure}[t]
   \begin{center}
      \includegraphics[width=1.0\linewidth]{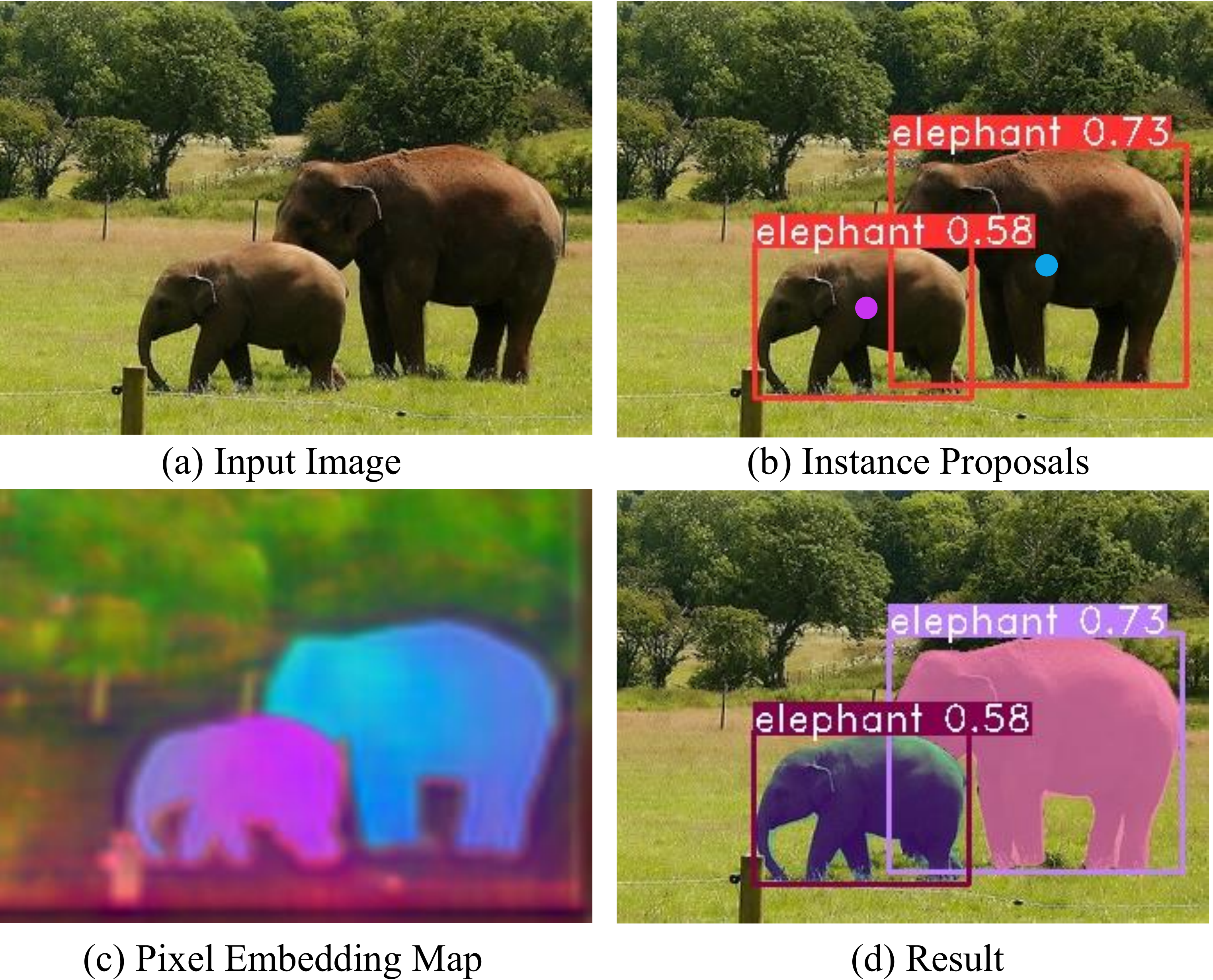}
   \end{center}
   \caption{(a) is an input image. (b) is the output instance proposals, attached with the parameters of bounding boxes, class scores, and proposal embeddings encoded by different colors. (c) is the output of the pixel embedding map encoded by different colors. (d) is the final result conducted from (b) and (c).
    For each proposal, pixels in the proposal box which have similar embeddings to the proposal embedding will be assigned to the mask of the proposal. }
   \label{fig:framework}
\end{figure}

There have been a variety of methods trying to solve this problem.
Proposal-based methods~\cite{chen2018masklab,dai2016instance_iccv,he2017mask} treat instance segmentation as an extension of object detection. When detected instances are determined with their bounding boxes, 
the segmentation task can be processed inside the box of each instance. 
As a representative, Mask R-CNN~\cite{he2017mask} achieves outstanding results on many benchmarks to be the most popular method for instance segmentation. 
However, as a two-stage method, the ``RoIPooling/RoIAlign'' step results in the loss of features and the distortion to the aspect ratios, so that the masks it produces may not preserve fine details. 
Besides, it still sustains weakness in being complex to adjust too many parameters. 
Recently, one-stage instance segmentation methods have become a popular topic, but those newly proposed methods cannot yet perform comparably to the two-stage ones. 
As one type of the one-stage method, segmentation-based methods~\cite{bai2017deep,de2017semantic,fathi2017semantic,kirillov2017instancecut,kong2018recurrent,liang2017proposal,neven2019instance} prefer to process the image in pixel-level directly so that they do not suffer from repooling operations. 
They predict features for each pixel and then the clustering process is applied to group them up for each object instance. 
However, the bottlenecks of such methods are their clustering procedures, such as the difficulties in determining the number of clusters or the positions of the cluster centers, resulting in the incomparable performance with the proposal-based methods. 

\paragraph{Our Contribution} In this work, we propose a new instance segmentation method which aims at exploiting the advantage of both proposal-based and segmentation-based methods. 
It preserves strong detection capabilities as the proposal-based methods, and meanwhile keeps the details of images as the segmentation-based methods.
In this way it is able to not only reach top scores in benchmark but also produce high-resolution masks and run at a high speed. 

Our method, named EmbedMask, is a one-stage method that for the first time achieves comparable results as Mask R-CNN in the challenging dataset COCO with the same training settings. 
Fundamentally, EmbedMask follows the framework of one-stage detection methods that it predicts instance proposals, which are defined by their bounding boxes, categories, and scores. 
As the key of segmentation-based methods, embedding is also used in our method for clustering, which we separate the embedding into a couple definitions: (1) embedding for pixels, referred as \emph{pixel embedding}, which is a representation for every pixel in the image, as shown in Figure~\ref{fig:framework}(c), and (2) embedding for instance proposals, referred as \emph{proposal embedding}, which is a representation for the instance proposals besides the bounding box and classification, as shown in Figure~\ref{fig:framework}(b). 
Embedding coupling is applied to the above embeddings that pixel embedding is supervised to couple with the proposal embedding if they correspond to the same instance.

In the inference procedure, each instance proposal surviving from the non-maximum suppression (NMS) is attached with a proposal embedding which is regarded as the cluster center that guides the clustering among pixel embeddings to generate the mask for the instance. 
With this process, we not only avoid determining the cluster centers as well as their number but remove the need for the computation of ``RoIPooling/RoIAlign''. 
In this way, we can keep the essential details while omitting the complex operations.
Furthermore, we predict another parameter for the instance proposal to produce certain margin for the clustering procedure, which is proposal-sensitive. 
The flexible margins make it more suitable to conduct instance segmentation for multi-scale objects, which is a mechanism that most of the one-stage methods do not possess.

EmbedMask simplifies the clustering procedure in the segmentation-based methods and avoid the repooling procedure in Mask R-CNN. 
While being simple but effective, our method produces a significant boost, compared to other contemporary work~\cite{bolya-iccv2019, xie2019polarmask}. 
Notably, EmbedMask achieves comparable results to Mask R-CNN, with the mask mAP of 37.7 vs. 38.1 in the challenging COCO dataset~\cite{lin2014microsoft} and speed of 13.7 fps vs. 8.7 fps (V100 GPU), both using the ResNet-101~\cite{he2016deep} as backbone network and under the same training settings. In summary, the main contributions of our work are mainly twofolds:
\begin{itemize}

\item We propose a framework that unites the proposal-based and segmentation-based methods, by introducing the concepts of proposal embedding and pixel embedding so that pixels are assigned to instance proposals according to their embedding similarity.
\item As a one-stage instance segmentation method, our method can achieve comparable scores as Mask R-CNN in the COCO benchmark, and meanwhile it provides masks with a higher quality than Mask R-CNN, running at a higher speed. 

\end{itemize}

\begin{figure*}
   \begin{center}
      \includegraphics[width=1.0\linewidth]{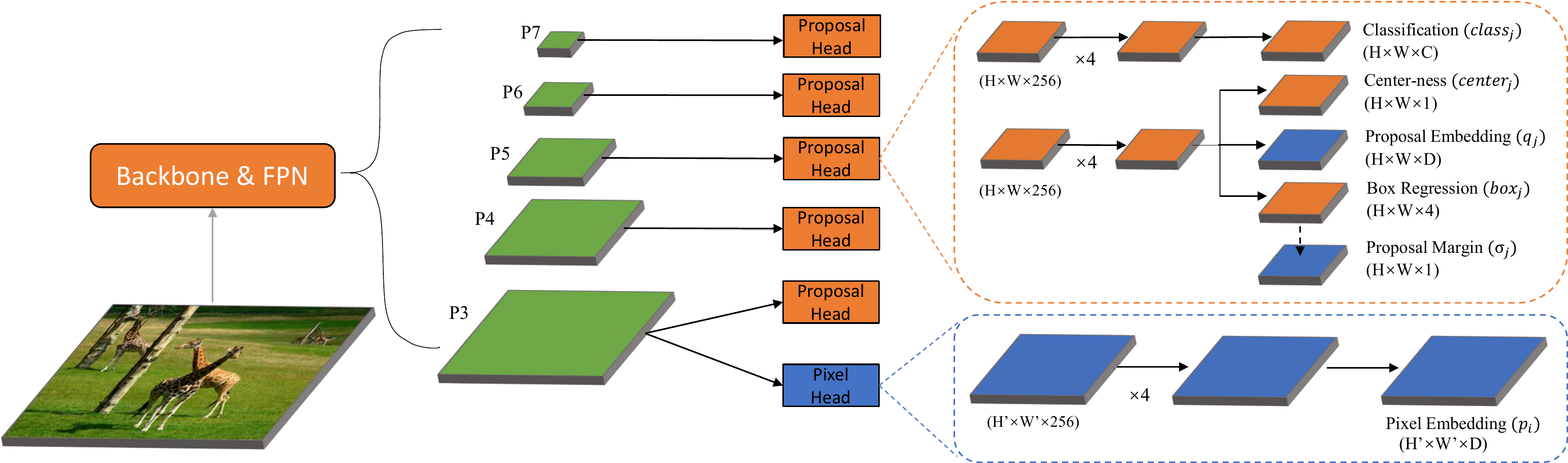}
   \end{center}
      \caption{EmbedMask shares most parts of network architecture with the 
      FCOS~\cite{tian2019fcos}. All the blue feature maps are newly added base on FCOS.
      In proposal head and pixel head, solid arrows indicate $3\times3\ conv$ layers and dotted arrow indicates $1\times1\ conv$ layer. 
      The $\times 4$ marks indicate feature maps pass 4 $3\times3\ conv$ layers.
      }
   \label{fig:architecture}
\end{figure*}

\section{Related Work}

Instance segmentation is a fundamental yet challenging task, which requires to predict a pixel-level mask with a category label for each instance of interest in an image. 
Various methods with different ideas have been proposed to solve this problem. 

\medbreak
\noindent {\bf Two-stage Methods} 
Two-stage methods can be thought of consisting of two consecutive stages: detection and segmentation, where the segmentation results depend on the detection results, as the segmentation is processed on each detected bounding box. 
Before the rise of the unified framework, Pinheiro \etal~\cite{pinheiro2015learning} proposed DeepMask, which utilizes sliding windows to generate proposal regions, and then learns to classify and segment them.
Mask R-CNN~\cite{he2017mask} unites the tasks of region proposing and segmentation built on top of Faster R-CNN~\cite{ren2015faster}, making it the representative of two-stage instance segmentation methods. 
Based on Mask R-CNN, PANet~\cite{liu2018path} enhances the performance by merging multi-level information. MS R-CNN~\cite{huang2019mask} simply redefines the grading standard of instance mask. 
With the detection models built on top of FPN~\cite{lin2017feature} as the baseline, recent two-stage instance segmentation methods achieve state-of-the-art performance. 
However, there still some remain problems, such as the low speed and detail-missing masks of large objects due to the complicate network architectures and the repooling step.

\medbreak
\noindent {\bf One-stage Methods}
Compared to the two-stage methods, the one-stage instance segmentation methods remove the repooling step. 
To avoid re-extracting features for instance proposals, \cite{dai2016instance_eccv,li2017fully} generate position-sensitive mask maps that can be simply assembled to get the final masks. 
TensorMask~\cite{chen2019tensormask} regard the instance segmentation task as an object detection task, and it replaces the 3D tensors for representing the bounding boxes to the 4D tensors to represent the masks over 2D spatial domain.
YOLACT~\cite{bolya-iccv2019} proposed the concept of prototype masks that can be linearly combined to generate instance masks. 
Though these methods have simpler procedures than the two-stage ones, they cannot produce masks that are accurate enough as the two-stage methods. 

\medbreak
\noindent {\bf Segmentation-based Methods} 
The segmentation-based methods are another kind of one-stage methods which divides the instance segmentation task to first segmenting and then clustering. 
Pixel-level predictions are obtained by the segmentation module and the clustering is applied to group them for each object.
For the separation of pixels on different objects and clustering of pixels on the same objects, \cite{de2017semantic} utilizes the discriminative loss while \cite{neven2019instance} introduces a new loss function that learns margins extra for different objects. 
Such bottom-up methods~\cite{de2017semantic, fathi2017semantic, neven2019instance} can naturally fetch high-resolution masks, but their performance in perceiving instances is not high enough.

Our proposed EmbedMask, as a one-stage method, can achieve comparable results with Mask R-CNN, and outperforms the state-of-the-art one-stage methods. 
Specifically, EmbedMask is built on top of the one-stage object detection method like other proposal-based instance segmentation methods~\cite{chen2018masklab,he2017mask}. 
As a key modification, we design a mask prediction module based on the \emph{proposal embedding} and \emph{pixel embedding} to perform efficient pixel clustering. 
Similar to the segmentation-based methods~\cite{de2017semantic,neven2019instance}, pixel clustering is performed on a predicted embedding map for the whole image so that EmbedMask can fetch high-resolution masks. 

\section{EmbedMask}

\subsection{Overview}

Our instance segmentation framework is composed of two parallel modules, one for finding the positions of instance proposals, and the other for predicting the masks of instance proposals. 
In practice, we use the state-of-the-art object detection method FCOS~\cite{tian2019fcos} as our baseline, which is the most recent one-stage object detection method. We note that our method can also be applied on other detection frameworks as~\cite{lin2017focal,liu2016ssd,redmon2016you}.

As the key of our method, we specially design extra modules to learn the pixel embeddings, proposal embeddings, and proposal margins to extract the instance masks.
Specifically, as shown in Figure~\ref{fig:architecture}, the three parameters are predicted from networks. 
First, the pixel embedding variants, referred as $p$, are computed in an additional single branch ``Pixel Head'' originating from the largest feature map of FPN, \ie, P3, with five $3\times3\ conv$ layers.
Second, the proposal embedding variants, referred as $q$, are computed by a $3\times3\ conv$ layer added after the feature map from FPN with another 4 $3\times3\ conv$ layers, which is shared with the prediction of center-ness and box regression.
Third, the proposal margin variants, referred as $\sigma$, are computed by a $1\times1\ conv$ layer added after the box regression outputs. 
All the predicted feature maps in the ``Proposal Head'' are united to produce the proposal features. That is, the values at the same location $x_j$ of these feature maps are grouped as a tuple $\{class_j, box_j, center_j, q_j, \sigma_j\}$ that represents the parameters of the $proposal_j$.

For each instance proposal and each pixel in the image, the distance between the couple of proposal embedding and the pixel embedding decides how likely this pixel belongs to the mask of the instance proposal, and the proposal margin gives a clear boundary for this likelihood to decide the final mask.

\subsection{Embedding Definition} \label{sec:embed_def}

As we know, the main task in instance segmentation is to assign  the pixels $x_i$ in the image to a set of instances $S_k$. 
The previous segmentation-based methods directly do this assignment by clustering pixels with similar embeddings, while we propose two new definitions to the embeddings, which are \emph{pixel embedding} and \emph{proposal embedding}. 
The proposal embedding represents the object-level context features for the object instance, which is a good representation of entire instance, while the pixel embedding represents the pixel-level context features for each location on the image, which learns the relation between each pixel with corresponding instance.
Proposal embeddings are used as cluster centers of instances to do pixel clustering among the pixel embeddings, so that the difficulties appeared in segmentation-based methods, such as finding the locations and counts of cluster centers, are avoided. 

Specifically, during inference, proposal embeddings and pixel embeddings are utilized for mask generation. 
In detail, after the NMS applied to the tuples of $\{box_j, class_j, center_j, q_j, \sigma_j\}$, a group of surviving instance proposals $S_k$ are fetched with these tuples as parameters. 
Here we name the corresponding $q_j$ for the surviving proposal $S_k$ as $Q_k$.
With the pixel embedding $p_i$ for each pixel $x_i$ in the image, a pixel $x_i$ is assigned to the instance proposal $S_k$ if the distance between the pixel embedding $p_i$ and the proposal embedding $Q_k$ are close enough. 
If we fix a margin $\delta$ to the distance, at inference time, the binary mask of $S_k$ can be computed by the pixel assignment

\begin{equation}
   Mask_k(x_i)=\begin{cases}
   1,\quad \left\|p_i-Q_k\right\|\leq \delta \\
   0,\quad \left\|p_i-Q_k\right\|>\delta.
   \end{cases}
   \label{eq:mask}
\end{equation}

During training, different from the inference time, $S_k$ is used to refer to each ground-truth instance, and the $Q_k$, being the proposal embeddings for the ground-truth instance $S_k$, now is the average of positive proposal embeddings. 
The positive proposal embeddings sampling strategy is described in section~\ref{sec:training}. 
Therefore, our objective is to bring the pixel embedding $p_i$ and proposal embedding $Q_k$ closer if the pixels $x_i$ belongs to the ground-truth mask of the instance $S_k$, otherwise keep them away. 

To perform such push and pull strategy for the foreground and background pixel embeddings, an intuitive method is to apply two fixed margins to two hinge losses, as 
\begin{equation}
\begin{aligned}
   L_{hinge} = 
   \frac{1}{K}\sum_{k=1}^{K}
   {
      \frac{1}{N_k}\sum_{i\in{\mathcal{B}_k}}
      {
         \mathbbm{1}_{\{i\in S_k\}} \left[\left\|p_i-Q_k\right\|-\delta_a\right]_+^2 
      }
   }\\
   +
   \frac{1}{K}\sum_{k=1}^{K}
   {
      \frac{1}{N_k}\sum_{i\in{\mathcal{B}_k}}
      {
         \mathbbm{1}_{\{i\notin S_k\}} \left[\delta_b-\left\|p_i-Q_k\right\|\right]_+^2.
      }
   }
   \label{eq:hinge_loss}
\end{aligned}
\end{equation}
In this function, $K$ is the number of ground-truth instances.
$\mathcal{B}_k$ represents the set of pixel embeddings that need to be supervised for the instance $S_k$, which is just the pixel embeddings locating inside the bounding box of $S_k$, and $N_k$ is the number of pixel embeddings in $\mathcal{B}_k$. 
$\mathbbm{1}_{\{i\in S_k\}}$ is an indicator function, being $1$ if pixel $x_i$ is in the ground-truth mask of $S_k$ and $0$ otherwise. 
$[x]_+=max(0, x)$, and the $\delta_a$ and $\delta_b$ are two margins designed for push and pull strategy. 
Specifically, the first term of the loss means to pull the distance between pixel embedding $p_i$ and proposal embedding $Q_k$ inside the margin $\delta_a$, and the second term means to push the distance outside the margin $\delta_b$.

However, we observe that such fixed margins may cause certain problems (see section~\ref{sec:learn_margin}), therefore we propose learnable margins to replace the fixed margins, which is more advantageous.

\subsection{Learnable Margin} \label{sec:learn_margin}

The loss function introduced above gives a solution to optimize the distance between pixel embeddings and proposal embeddings during training. 
However it uses the fixed margins for all instances, which may lead to some problems in training. 
First, the margins $\delta_b$ and $\delta_a$, as well as $\delta$ for inference, all need to be set manually, so it is difficult to find the optimal values for the best performance. 
Second, fixed margins for all instances are not friendly to the training of multi-scale objects, as the pixel embeddings in large objects are always more scattered while those in small objects are always concentrated.
In order to get rid of the problem, we propose the margins $\sigma_j$ for all instance proposals, which are flexible with multi-scale objects. 
Moreover, the flexible margins $\sigma_j$ can be learned directly from the training without the manual setting. 

To reach this point, inspired by~\cite{neven2019instance}, we use a Gaussian function, as
\begin{equation}
   \phi{\left(x_i, S_k\right)} = 
   \phi{\left(p_i, Q_k, \Sigma_k\right)} = 
   \exp{\left(-\frac{
      \left\|p_i-Q_k\right\|^2
   }{
      2\Sigma^2_k
   }\right)},
   \label{eq:probability}
\end{equation}
to map the distance between the pixel embedding $p_i$ of the pixel $x_i$ and the proposal embedding $Q_k$ of the instance $S_k$ into a value ranged in $[0, 1)$. 
The additional introduced variant $\Sigma_k$ comes from $\sigma_j$ just like how $Q_k$ comes from $q_j$.

The $\phi{\left(x_i, S_k\right)}$ is the probability for the pixel $x_i$ belonging to the mask of the instance $S_k$. When the pixel embedding $p_i$ is close to the proposal embedding $Q_k$, $\phi{\left(x_i, S_k\right)}$ is going to be $1$, otherwise $0$.
As what is introduced in~\cite{neven2019instance}, the $\Sigma_k$ plays a role of margin for instance $S_k$. 
So that in our method, the predicted $\sigma_j$ gives the learnable margin for each instance proposal. 

For the instance $S_k$, when the $\phi{\left(x_i, S_k\right)}$ is applied to each pixel $x_i$ in the image, a foreground/background probability map for the instance is produced. Therefore it can be optimized by a binary classification loss, which is 
\begin{equation}
   L_{mask} = 
   \frac{1}{K}\sum_{k=1}^{K}
   {
      \frac{1}{N_k}\sum_{p_i\in{\mathcal{B}_k}}
      {
         \mathcal{L}\left({
            \phi{\left(x_i, S_k\right)}, 
            \mathbb{G}(x_i, S_k)
         }\right),
      }
   }
   \label{eq:mask_loss}
\end{equation}
where $\mathcal{L}(\cdot)$ is a binary classification loss function, and in practice we use lovasz-hinge loss~\cite{yu2015learning} for better performance. 
$\mathbb{G}(x_i, S_k)$ represents the ground truth label for pixel $x_i$ to judge whether it is in the mask of the proposal $S_k$, which is a binary value.

This loss function supervises the computed mask probability maps, which contains the parameters of pixel embedding $p$, proposal embedding $q$ and proposal margin $\sigma$. 
So that the proposal margins can be learned automatically without manual settings.
And the flexible margin for each instance makes it more advantageous than the hinge loss. 

\subsection{Smooth Loss}

As mentioned above, the meanings of $Q_k$ and $\Sigma_k$ are with subtle difference in training and inference.

During training, $S_k$ represents the ground-truth object instance. For each instance $S_k$, the computation of $Q_k$ and $\Sigma_k$ is by averaging a set of positive samples $q_j$ and $\sigma_j$, and we name this set as $\mathcal{M}_k$ (described in section~\ref{sec:training}).
Specifically, the $Q_k$ and $\Sigma_k$ are computed as 
\begin{equation}
   Q_k = \frac{1}{N_k}\sum_{j\in{\mathcal{M}_k}}{q_j},
   \label{eq:avg_q}
\end{equation}
\begin{equation}
   \Sigma_k = \frac{1}{N_k}\sum_{j\in{\mathcal{M}_k}}{\sigma_j},
   \label{eq:avg_sigma}
\end{equation}
where $N_k$ is the number of positive samples of $S_k$ for proposal embedding and margin. 

But in the inference procedure, $S_k$ represents each instance proposal surviving from NMS. 
And the corresponding $q_j$ and $\sigma_j$ for surviving $S_k$ are used as $Q_k$ and $\Sigma_k$.

Because the $Q_k$ and $\Sigma_k$ are different when training and inference, 
we need to add a smooth loss for training to force them keeping close as

\begin{equation}
\begin{aligned}
   L_{smooth} = &\frac{1}{K}\sum_{k=1}^{K}{
       \frac{1}{N_k}\sum_{j\in{\mathcal{M}_k}}{
          \left\|q_j-Q_k\right\|^2
       }
   }\\
   +&\frac{1}{K}\sum_{k=1}^{K}{
        \frac{1}{N_k}\sum_{j\in{\mathcal{M}_k}}{
            \left\|\sigma_j-\Sigma_k\right\|^2.
        } 
   }
   \label{eq:smooth_loss}
\end{aligned}
\end{equation}

\begin{table*}
   \setlength{\abovecaptionskip}{-15pt}
   \setlength{\belowcaptionskip}{0pt}
   \begin{center}
   \begin{spacing}{1.0}
   \begin{threeparttable}
   \begin{tabular}{c|c|c|c|c|
      c c c|
      c c c |
      c | c c}
   method & backbone & ms & rc & epochs & 
   AP & $\text{AP}_{50}$ & $\text{AP}_{75}$ & 
   $\text{AP}_S$ & $\text{AP}_M$ & $\text{AP}_L$ & 
   $\text{AP}^{bb}$ & fps \\
   \hline
   Mask R-CNN,ours & R-50-FPN &  & & 12 & 
   34.6 & 56.5 & 36.6 & 15.3 & 36.3 & 49.7 & 38.0 & 8.6 \\
   Mask R-CNN,ours & R-101-FPN & & & 12 & 
   36.2 & 58.6 & 38.5 & 16.4 & 38.4 & 52.0 & 40.1 & 8.1 \\
   Mask R-CNN,ours & R-101-FPN & $\checkmark$ & & 36 & 
   38.1 & 60.9 & 40.7 & 18.4 & 40.2 & 53.4 & 42.6 & 8.7 \\
   PANet~\cite{liu2018path} & R-50-FPN & \checkmark &  & 22 & 
   \textbf{38.2} & 60.2 & 41.4 & 19.1 & 41.1 & 52.6 & - & 4.7 \\
   RetinaMask\cite{fu2019retinamask} & R-101-FPN& $\checkmark$ &  & 24 & 
   34.7 & 55.4 & 36.9 & 14.3 & 36.7 & 50.5 & 39.1 & 6.0 \\
   \hline
   TensorMask~\cite{chen2019tensormask} & R-101-FPN & $\checkmark$ & & 72 & 
   37.3 & 59.5 & 39.5 & 17.5 & 39.3 & 51.6 & 41.6  & 2.6 \\
   YOLACT-700\cite{bolya-iccv2019} & R-101-FPN & $\checkmark$ & $\checkmark$ & 48 & 
   31.2 & 50.6 & 32.8 & 12.1 & 33.3 & 47.1 & -  & 23.6 \\
   EmbedMask & R-50-FPN  & & & 12 & 
   33.6 & 54.5 & 35.4 & 15.1 & 35.9 & 47.3 & 38.2 & 16.7 \\
   EmbedMask & R-101-FPN & & & 12 & 
   35.6 & 56.8 & 38.0 & 16.2 & 38.1 & 50.6 & 40.2 & 13.5 \\
   EmbedMask & R-101-FPN & $\checkmark$ & & 36 & 
   \textbf{37.7} & 59.1 & 40.3 & 17.9 & 40.4 & 53.0 & 42.5 & 13.7\\
   EmbedMask-600 & R-101-FPN & $\checkmark$ &  & 36 & 
   35.2 & 55.9 & 37.3 & 12.4 & 37.3 & 54.9 & 40.2 & 21.7 \\
   \end{tabular}
   \end{threeparttable}
   \end{spacing}
   \end{center}
   \caption{Comparison with state-of-the-art methods for instance segmentation on COCO \textit{test-dev}. 
   The methods located above are two-stage ones, and below are one-stage. 
   In the table, `ms' and `rc' means multi-scale and random crop for training. 
   `EmbedMask-600'uses the same trained model as `EmbedMask', while doing inference with the smaller input images whose shorter sides are 600 and longer sides are no longer than 800.}
   \label{tab:mask_performance}
\end{table*}

\subsection{Training} \label{sec:training}

\medbreak
\noindent {\bf Objective} 
EmbedMask is optimized end-to-end using a multi-task loss. Apart from the original classification loss $L_{cls}$, center-ness loss $L_{center}$ and box regression loss $L_{box}$ in FCOS, we introduce additional losses $L_{mask}$ and $L_{smooth}$ for mask prediction. 
They are jointly optimized by 

\begin{equation}
   L=L_{cls} + L_{center} + L_{box} + \lambda_1L_{mask} + \lambda_2L_{smooth}.
   \label{eq:loss}
\end{equation}

\medbreak
\noindent {\bf Training Samples for Box and Classification}
When computing the losses for box regression and classification, as well as center-ness, we define the positive samples as the parameters $\{box_j, class_j, center_j\}$ whose real locations mapped back to the original image locate on the center region of the ground-truth bounding box, and at the meantime the locations are in the mask of the ground-truth instances. 
The sampling strategy is a little more strict than the original one in FCOS, that we enforce the sample to be more accurate to the mask-level.

\medbreak
\noindent {\bf Training Samples for Proposal Embedding and Margin}
Compared to the sampling for box and classification, the sampling of proposal embeddings $q_j$ and margins $\sigma_j$, which are used to compute $Q_k$ and $\Sigma_k$ for training, need another condition, that the Intersection over Union (IoU) between the corresponding predicted $box_j$ in the sampled location and the ground-truth box for instance $S_k$ should be more than $0.5$. 
This more strict selection strategy reduces positive samples, so that also reduces the training difficulty.

\noindent {\bf Training Samples for Pixel Embedding}
In the definition of mask loss, as shown in Equation~\ref{eq:probability}, only the pixels belong to $\mathcal{B}_k$ are supervised for the instance $S_k$. 
In our experiment, the $\mathcal{B}_k$ is the set of samples that lay inside the ground-truth bounding box of $S_k$ when training. 
But in practice, we find that if we slightly expand the box to increase the number of training samples, the results will be better. 
Thus, the manual expand to the box is used in our method.

\subsection{Inference} 
The inference procedure of EmbedMask is very clear. 
Given an input image, it will go through the object detection procedure as FCOS, and the instances that survive NMS are treated as instance proposals $S_k$. 
Each surviving proposal $S_k$ is attached with its bounding box, the category with a related score, the proposal embedding $q_j$, and the proposal margin $\sigma_j$. 
The $q_j$ and $\sigma_j$ are just viewed as $Q_k$ and $\Sigma_k$ in inference. 
In the meantime, we can also obtain the pixel embedding $p_i$ for each pixel $x_i$ in the image. 
For each pixel $x_i$ in the bounding box of $S_k$, we can use Equation~\ref{eq:probability} to calculate the probability of $x_i$ belong to $S_k$, then translate the probability to binary value using a thresh 0.5. In this way, the final masks are produced. 

\section{Experiments}

\subsection{Training Details}

\medbreak
\noindent {\bf Training Data}
We follow the settings of FCOS in our experiments, which chooses the large-scale detection benchmark COCO, and uses the COCO \textit{trainval35k} split (115K images) for training, \textit{minival} split (5K images) for ablation study and \textit{test-dev} (20K images) for reporting the main results. 
The input images are resized with the short side being 800 while the longer side being no longer than 1333.

\medbreak
\noindent {\bf Network Architecture}
ResNet-50~\cite{he2016deep} is used as our backbone network for ablation study; ResNet-101 is used for comparing results with state-of-the-art methods. 
Following the FPN architecture as FCOS, we also use five levels of feature maps defined as $\{P3,P4,P5,P6,P7\}$.
The architecture for outputting proposal margins is a little different, as it actually predicts $\frac{1}{2\sigma_j^2}$ instead of $\sigma_j$ directly, so the variant is activated by the exponential function to keep it positive. 

\medbreak
\noindent {\bf Training and Inference Procedure}
We train all the models with SGD for $90k$ iterations using an initial learning rate of $0.01$ and batch size of $16$, with constant warm-up of 500 iterations. 
The backbone network is initialized with the pre-trained ImageNet~\cite{deng2009imagenet} weights. 
In default we set $\lambda_1=0.5$, $\lambda_2=0.1$ and embedding dim $d=32$. 
For the main results in Table~\ref{tab:mask_performance}, the box expand with $1.2\times$ is used in producing training samples $\mathcal{B}_k$ for pixel embedding (while box expand with $1.0\times$ for ablation study by default).
For the alignment of $\phi{\left(x_i, S_k\right)}$ and $\mathbb{G}(x_i, S_k)$ in the mask loss~\ref{eq:mask_loss} during training, we resize the feature map of pixel embedding and the ground-truth mask for each instance to be a quarter of the input image in length, using  bilinear interpolation. 
During inference, we do the same for the feature map of pixel embedding and then re-scale it to the initial size to obtain the mask.

\begin{figure*}
   \begin{center}
      \includegraphics[width=1.0\linewidth]{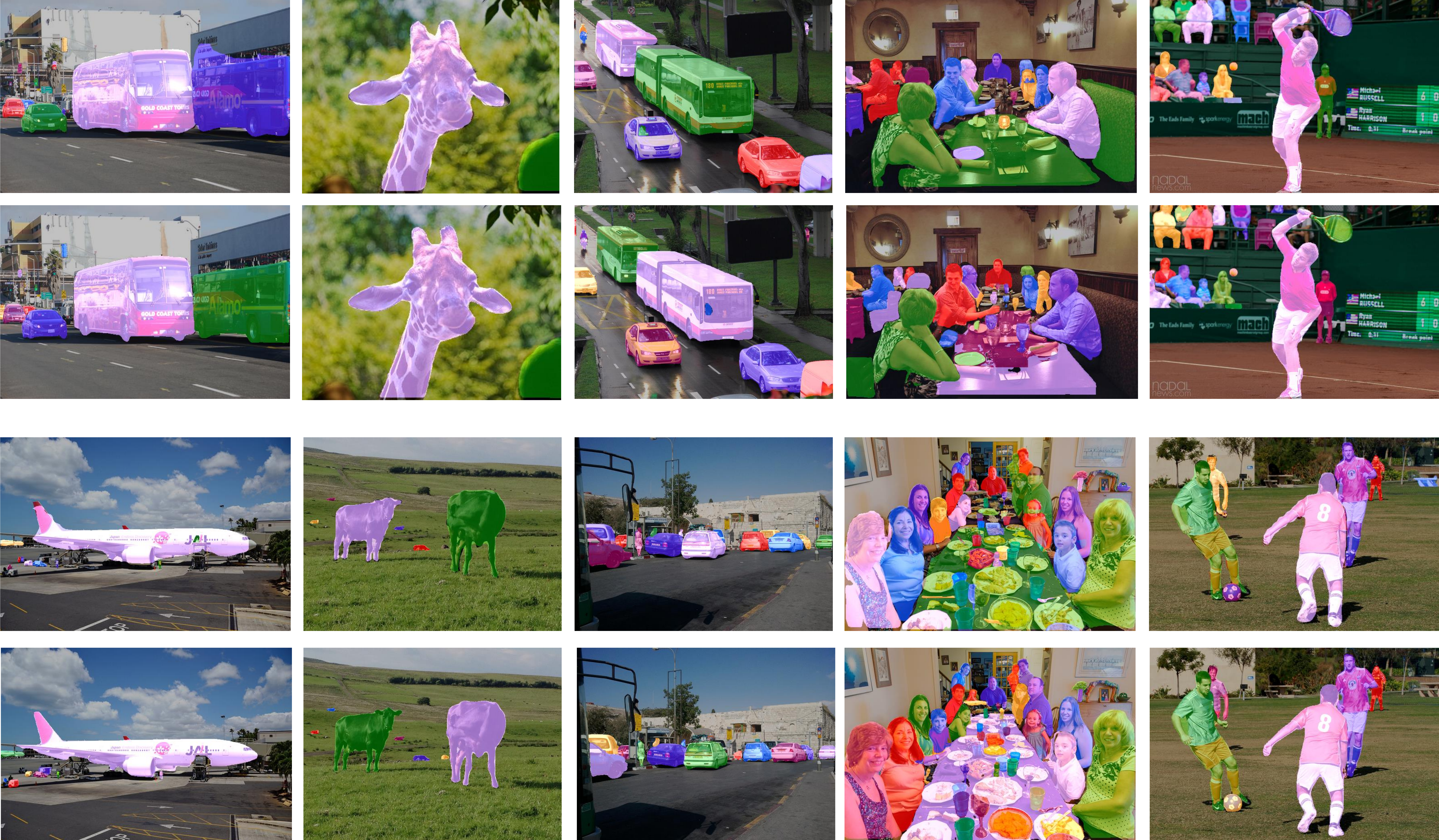}
   \end{center}\vspace{-0.1in}
      \caption{Visualization results. The upsides are the masks of Mask R-CNN and the downsides are the masks of EmbedMask, both with the ResNet-101 backbone and under the same training settings.}
   \label{fig:vis}
\end{figure*}

\subsection{Main Results}

\medbreak
\noindent {\bf Quantitative Results}
We compare the quantitative results of EmbedMask with other state-of-the-arts methods, including one-stage and two-stage methods, which are shown in Table~\ref{tab:mask_performance}. 
The listed results are all trained with ResNet-50-FPN or ResNet-101-FPN as the backbone for fairness. 
It is worth noting that some of the one-stage methods applied several training tricks for better performance, such as more training epochs. 
Hence it is not fair to directly compare EmbedMask with those methods. 
Nevertheless, even without data augmentation or more training epochs, our method outperforms YOLACT with $4.4$ AP. 
And with more training epochs and data augmentation, our method has better performance and faster speed than TensorMask, and achieves the best performance within one-stage methods. 
With the scale of 600, `EmbedMask-600' can achieve faster speed but the accuracy decreases because the scale is not matched with the one in training procedure. 
From the table, we can find that `EmbedMask-600' has similar speed as YOLACT but better performance. 

We also focus on the comparison between Mask R-CNN and our method, which both use the same training settings. 
The gap of mask AP between our method and Mask R-CNN is about $0.4$ in ResNet-101 with multi-scale training, which is quite close. 
We also observe that the gap between EmbedMask and Mask R-CNN in ResNet-101 is $1.8$ in $\text{AP}_{50}$, but $0.4$ in $\text{AP}_{75}$. 
That means our method is more advantageous in providing more accurate masks. 
Additionally, in comparison with the speed of 8.6 fps in Mask R-CNN, our method can run in 16.7 fps with 33.6 mAP for an input image both use the backbone of ResNet-50 and with the short side being 800 on a V100 GPU. 
As for the speed in ResNet-101, our method runs in 13.7 fps while Mask R-CNN runs in 8.7 fps. 
Therefore, about running speed, our method runs much faster than Mask R-CNN.
The implementation of EmbedMask and Mask R-CNN are all based on maskrcnn-benchmark~\cite{massa2018mrcnn}, and the inference time of the both includes the forwarding time of the networks and the postprocessing time including mask postprocessing which restores the masks to the sizes of input images.

\begin{figure*}
   \begin{center}
      \includegraphics[width=1.0\linewidth]{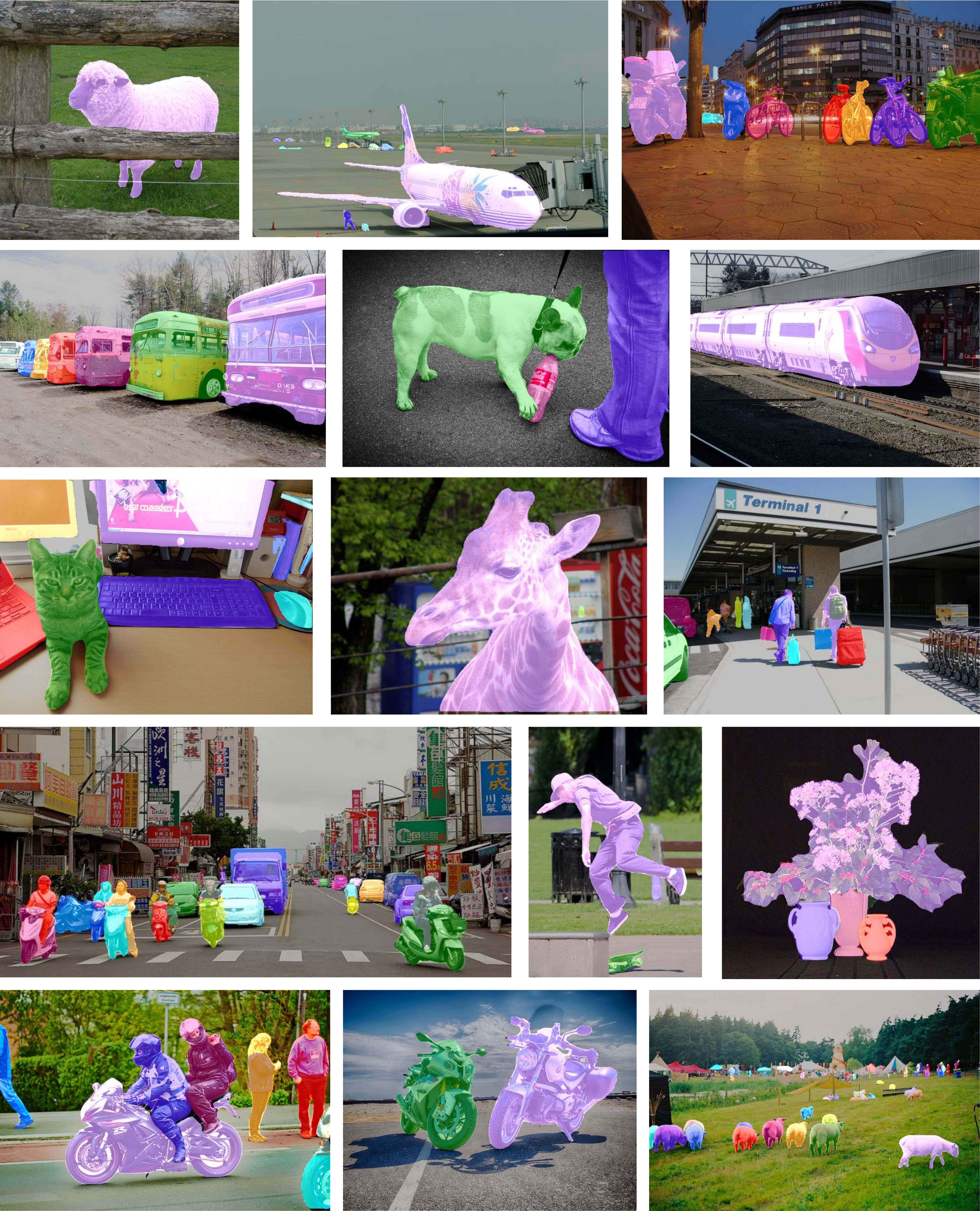}
   \end{center}\vspace{-0.1in}
      \caption{Mask visualizations}
   \label{fig:maskvis}
\end{figure*}

\medbreak
\noindent {\bf Qualitative Results}
Similar to other segmentation-based methods, EmbedMask generates masks directly from the output feature maps without the need of repooling. 
Specifically, in EmbedMask, the masks are generated from the feature map of pixel embedding which is predicted directly from the largest feature map of FPN with a stride 8, which can produce more detail masks. 
Figure~\ref{fig:vis} visualizes the comparison of the mask quality between the Mask R-CNN and EmbedMask, and both of these results are from models trained with 12 epochs and without multi-scale. 
As we can see from the qualitative results, our method can provide more detailed masks than Mask R-CNN with sharper edges, and that is because our method do not use the repooling operation so that avoid missing details. 
More visualization results which come from the model trained with multi-scale and 36 epochs can be found in Figure~\ref{fig:maskvis}. 

\begin{table*}[t]
   \begin{center}
   \subcaptionbox{
      \textbf{Fixed vs. learnable margin.} Results of using fixes margin and learnable margin. 
      \label{subcap:margin}
   }[0.48\linewidth]
   {
      \begin{tabular}{p{0.9cm}<{\centering}|p{0.7cm}<{\centering}|
         p{0.55cm}<{\centering} p{0.55cm}<{\centering} p{0.55cm}<{\centering}|
         p{0.55cm}<{\centering} p{0.55cm}<{\centering} p{0.55cm}<{\centering}}
         loss & $\sigma$ &
         AP & $\text{AP}_{50}$ & $\text{AP}_{75}$ & 
         $\text{AP}_S$ & $\text{AP}_M$ & $\text{AP}_L$ \\
         \hline
         $L_{hinge}$ & - & 30.0 & 52.0 & 30.2 & 13.8 & 33.1 & 43.4 \\
         $L_{mask}$ & const & 33.1 & 53.8 & 34.7 & 14.7 & 36.4 & 47.9 \\
         $L_{mask}$ & pred & \textbf{33.3} & 54.0 & 35.0 & 15.8 & 36.6 & 47.7
      \end{tabular}
   }
   \hspace{0.02\linewidth}
   \subcaptionbox{
      \textbf{The choice of cluster center.} Results of using proposal embedding as cluster center and using pixel embedding as cluster center. 
      \label{subcap:proposal}
   }[0.48\linewidth]
   {
      \begin{tabular}{p{1.2cm}<{\centering}|
         p{0.55cm}<{\centering} p{0.55cm}<{\centering} p{0.55cm}<{\centering}|
         p{0.55cm}<{\centering} p{0.55cm}<{\centering} p{0.55cm}<{\centering}}
         $q_j$ &
         AP & $\text{AP}_{50}$ & $\text{AP}_{75}$ & 
         $\text{AP}_S$ & $\text{AP}_M$ & $\text{AP}_L$ \\
         \hline
         pixel  & 30.9 & 52.0 & 31.9 & 14.0 & 34.9 & 44.5 \\
         proposal & \textbf{33.3} & 54.0 & 35.0 & 15.8 & 36.6 & 47.7 \\
         \hline
         $\Delta$ & +2.4 & +2.0 & +3.1 & +1.8 & +1.7 & +3.2
      \end{tabular}
   }
    \medbreak
    \subcaptionbox{
      \textbf{Sampling strategy.} Results of using different sampling strategy, verifying that whether the samples should be in the ground-truth mask and with an IoU larger than 0.5.
      \label{subcap:sampling}
    }[0.34\linewidth]
    {
      \begin{tabular}{p{1.1cm}<{\centering}|p{1.2cm}<{\centering}|
         p{0.5cm}<{\centering} p{0.5cm}<{\centering} p{0.5cm}<{\centering}}
         in mask & IoU$>$0.5 &
         AP & $\text{AP}_{50}$ & $\text{AP}_{75}$ \\
         \hline
          & & 32.6 & 53.9 & 34.0  \\
         $\checkmark$ & & 32.9 & 53.7 & 34.3  \\
         $\checkmark$ & $\checkmark$ & \textbf{33.3} & 54.0 & 35.0 
      \end{tabular}
    }
    \hspace{0.01\linewidth}
    \subcaptionbox{
      \textbf{Training samples for pixel embedding.} 
      Results of using different number of training samples for pixel embeddings, which is controlled by the box size.
      \label{subcap:box}
    }[0.29\linewidth]
    {
      \begin{tabular}{p{1.0cm}<{\centering}|
         p{0.55cm}<{\centering} p{0.55cm}<{\centering} p{0.55cm}<{\centering}}
         box &
         AP & $\text{AP}_{50}$ & $\text{AP}_{75}$ \\
         \hline
         $1\times$ & 33.3 & 54.0 & 35.0   \\
         $1.2\times$ & \textbf{33.5} & 54.1 & 35.3  \\
         $1.5\times$ & 33.1 & 53.7 & 34.9
      \end{tabular}
    }
    \hspace{0.01\linewidth}
    \subcaptionbox{
      \textbf{Embedding dimension.} 
      Results of using different embedding dimension.
      \label{subcap:dim}
    }[0.29\linewidth]
    {
      \begin{tabular}{p{1.0cm}<{\centering}|
         p{0.55cm}<{\centering} p{0.55cm}<{\centering} p{0.55cm}<{\centering}}
         dim &
         AP & $\text{AP}_{50}$ & $\text{AP}_{75}$ \\
         \hline
         8 & 33.0 & 54.0 & 34.6   \\
         16 & 33.2 & 54.0 & 34.8  \\
         32 & \textbf{33.3} & 54.0 & 35.0
      \end{tabular}
    }
   \end{center}
   \vspace{-0.2in}
   \caption{Ablations on EmbedMask evaluated on COCO \textit{minival}. All models are training for 12 epochs with ResNet-50. \label{table:ablation_study}}
\end{table*}

\subsection{Ablation Study}

\medbreak
\noindent {\bf Fixed vs. Learnable Margin}
In the section \ref{sec:embed_def} and \ref{sec:learn_margin}, we introduce two kinds of loss functions for training. 
For the distance between proposal embedding and pixel embedding, the hinge loss function (Equation~\ref{eq:hinge_loss}) uses $\delta_a$ and $\delta_b$ to control the margins of foreground and background when training, and the margin $\delta$ is also fixed when training. 
While the mask loss function (Equation ~\ref{eq:mask_loss}) uses a gaussian function (Equation~\ref{eq:mask}) to map the distance to the probability, where the predicted $\Sigma_k$ is predicted to control the margin for each instance. 
But the $\Sigma_k$ can also be fixed as a constant for every instance. 
In the experiment, we set $\delta_a=0.5,\ \delta_b=1.5$ to test the performance of the hinge loss along with the fixed margins, and this leads to $\delta=0.8$ which shows best result in the inference phase. 
From the results in Table~\ref{subcap:margin}, we can find that the mask loss with the gaussian function outperforms the hinge loss a lot for avoiding the manual turning of parameters. And with the same mask loss, the flexible margins for each instance can perform better than the constant margin.

\medbreak
\noindent {\bf The Choice of Cluster Centers}
In EmbedMask, we use the embedding $Q_k$ as the cluster center of the proposal, which comes from the proposal embedding $q_j$. 
In fact, for each location $x_j$ on the feature map of proposal embedding, it can be mapped to the location $x_i$ on the feature map of pixel embedding with the embedding value $p_i$, so what if we replace the proposal embedding $q_j$ at the location $x_j$ with the pixel embedding $p_i$ at the mapped location $x_i$ and use this pixel embedding as the cluster center? 
The answer can be found in Table~\ref{subcap:proposal}. 
The replacement discards the need for predicting the proposal embedding, making the method more like a segmentation-based method. 
However, we can see that the additional predicted proposal embedding has a better performance. 
The credit may be given to the receptive field of proposal embedding, which is larger than that of pixel embedding and thus performs better.

\medbreak
\noindent {\bf Sampling Strategy}
As described in section \ref{sec:training}, during training, the positive samples for the $\{box_j, class_j, center_j\}$ requires the sampled location to be inside the mask of the ground-truth instance. 
While the positive samples for $\{q_j, \sigma_j\}$ needs an additional condition that the IoU between the predicted box $box_j$ and ground-truth box should be larger than 0.5. 
Here we discuss whether the sampling of $\{box_j, class_j, center_j, q_j, \sigma_j\}$ needs to be inside the ground-truth masks, and that of $\{q_j, \sigma_j\}$ need with the IoU greater than 0.5. 
The results can be found in Table~\ref{subcap:sampling}. 
We can find that the samples of the proposal embedding and margin constrained inside the ground-truth mask and with an IoU greater than 0.5 can obtain better results.

\medbreak
\noindent {\bf Training Samples for Pixel Embedding}
The training samples of pixel embedding for each ground-truth instance $S_k$, which are named $\mathcal{B}_k$, are all the samples located inside the bounding box of the $S_k$. 
The bounding box can be enlarged for more training samples. 
In the experiment described by Table~\ref{subcap:box}, we explore the suitable number of samples for training. 
The larger number of training samples with a $1.2\times$ bounding box in length can result in a better performance than the one with the original box. 
However, the performance drops when the bounding box gets larger. 
The experiment shows that we can select a suitable number of training samples to achieve the best results.

\medbreak
\noindent {\bf Embedding Dimension}
Table~\ref{subcap:dim} shows the results of different embedding dimensions. 
We can find that our method is robust with the embedding dimension. Even with a dimension of 8, the final mAP is still similar to that of 32.

\section{Conclusion}

In this work, we have proposed a single shot instance segmentation method, named EmbedMask, which can absorb the advantages of proposal-based and segmentation-based methods and avoid their weakness.
Specifically, EmbedMask makes use of the unique output feature map, named pixel embedding, to produce masks which can preserve fine details. 
Additionally, EmbedMask predicts the embedding for each instance proposal to cluster pixels in the feature map of pixel embedding according to their similarity. 
The predicted margins also adapt EmbedMask to the training for multi-scale objects.
In summary, as a single shot method, EmbedMask achieves comparable scores with the two-stage methods and run faster. 
In the future, we will improve EmbedMask by applying a more suitable network architecture to make it run faster and perform better. 
We hope our EmbedMask can inspire further research in the one-stage instance segmentation.


{\small
\bibliographystyle{ieee_fullname}
\bibliography{egbib}
}

\end{document}